\documentclass[9pt]{article} 
\usepackage{spconf,amsmath,graphicx,amssymb}

\usepackage{caption}
\usepackage{subcaption}
\usepackage{pgfplots}
\usepackage{standalone}

\usepackage{algorithm}
\usepackage{algpseudocode}
\usepackage{bbm}
\usepackage{graphicx}

\newcommand{\argmin}{\operatornamewithlimits{argmin}}

\algnewcommand\algorithmicinput{\textbf{Input:}}
\algnewcommand\INPUT{\item[\algorithmicinput]}
\algnewcommand\algorithmicoutput{\textbf{Output:}}
\algnewcommand\OUTPUT{\item[\algorithmicoutput]}

\title{A multi-layer network based on Sparse Ternary Codes\\for universal vector compression}
\name{Sohrab Ferdowsi, Slava Voloshynovskiy, Dimche Kostadinov}
\address{Dep. of Computer Science, University of Geneva, Switzerland\\
$\lbrace$sohrab.ferdowsi, svolos, dimche.kostadinov$\rbrace$@unige.ch
}

\begin{document}
%
\maketitle
\begin{abstract}
We present the multi-layer extension of the Sparse Ternary Codes (STC) for fast similarity search where we focus on the reconstruction of the database vectors from the ternary codes. To consider the trade-offs between the compactness of the STC and the quality of the reconstructed vectors, we study the rate-distortion behavior of these codes under different setups. We show that a single-layer code cannot achieve satisfactory results at high rates. Therefore, we extend the concept of STC to multiple layers and design the ML-STC, a codebook-free system that successively refines the reconstruction of the residuals of previous layers. While the ML-STC keeps the sparse ternary structure of the single-layer STC and hence is suitable for fast similarity search in large-scale databases, we show its superior rate-distortion performance on both model-based synthetic data and public large-scale databases, as compared to several binary hashing methods.
\end{abstract}
\section{Introduction}
\label{sec:intro}

The framework of Sparse Ternary Codes (STC) \cite{Sohrab_ISIT2017},  \cite{Sohrab_WIFS2016} was proposed as an alternative to binary hashing to address the problem of Approximate Nearest Neighbor (ANN) search. In ANN, compact representations should be designed to replace real-valued high-dimensional feature vectors in order to accommodate  large-scale data in small memory and perform fast search within them. The main idea behind the STC framework was to show that projecting the data to higher dimensions and quantizing them to a ternary but sparse alphabet should be preferred to projecting to lower dimensions and quantizing them to a dense binary alphabet since it provides better compactness and fidelity of representation. In particular, it was shown in \cite{Sohrab_ISIT2017} that for a fixed rate-budget, i.e., for a fixed amount of entropy of the codes, the STC framework preserves larger mutual information between the encoded versions of the data and its noisy realizations as compared to the popular binary hashing framework.

To perform more accurate similarity search, one can think of refining the inaccurate short-list returned by the compact codes by reconstructing the elements from the list and improve it by matching in the original feature domain. Motivated by this idea of ``list-refinement'', we focus in this paper on the reconstruction of the feature vectors from their STC by considering the rate-distortion trade-offs of the design.   

We provide the following contributions in this paper. First, complementary to the arguments of coding gain of \cite{Sohrab_ISIT2017} where it was shown that for a fixed compactness, ternary encoding provides higher mutual information between the encoded data and query, we show in this paper that they also provide better reconstruction quality compared to the binary codes for a fixed number of bits used.

Next, we show that in order to provide high-quality reconstruction, i.e., to operate at high rate regimes of the rate-distortion curve, one single layer of representation is not sufficient. Instead, we show that multi-layers of STC where each of the layers is operating within a very low-rate regime, or equivalently with highly sparse ternary values, is able to approximate the information-theoretic limits of the rate-distortion trade-off.\footnote{Since binary encoding is the special case of ternary encoding when there is no sparsity, this also implies that one cannot benefit from multi-layer design based on binary codes to achieve high-quality reconstruction.}

Finally, we test the rate-distortion performance of the system for different data sources. We first focus on the \textit{i.i.d.} Gaussian data and also correlated data with known covariance matrix with varying levels of correlation and consider the theoretical Shannon Lower Bound (SLB) as a guiding principle for design. We next show that we achieve very good compression performance on the MNIST set of images and GIST-1M set of descriptor features when compared to binary hashing methods from literature.

Section \ref{problem_formulation} formalizes the problem of compressive data-representation, particularly in view of fast search applications. After briefly reviewing the STC framework, section \ref{STC_Rec} focuses on the problem of  reconstruction from them. In section \ref{exp} we perform experiments on public databases and finally we conclude the paper in section \ref{summary}. 
\section{Problem Formulation: compressive data-representation for ANN search}
\label{problem_formulation}
Consider a database $\mathrm{F} = [\mathbf{f}(1), \cdots, \mathbf{f}(N)]$, consisting of  data-points $\mathbf{f}(i)$'s $\in \Re^n$, each of them features representing an entity like human biometrics, images or image descriptors. When either $N$ or $n$ is large, for a wide variety of tasks, it is crucial to store these vectors compressed. So we seek a compressive data representation scheme that should provide an encoder-decoder pair, $\mathbb{Q}[\cdot]$ and $\mathbb{Q}^{-1}[\cdot]$, such that the codes or the representations are as compact as possible. At the same time, when decoded, the codes should closely approximate the original data. The first requirement is characterized by rate and the second requirement by distortion.

More formally, for any realization vector $\mathbf{f}$, the rate of the representation $\mathbf{x} = \mathbb{Q}[\mathbf{f}]$ is defined as in (\ref{subeq:R}) and the distortion of the reconstruction $\hat{\mathbf{f}} = \mathbb{Q}^{-1}[\mathbf{x}]$ is defined as in (\ref{subeq:D})\footnote{We can think of the the random variable $\mathbf{F}$ (and likewise $\hat{\mathbf{F}}$) as the underlying rule that generates $\mathbf{f}$'s as its realizations. If $p(\mathbf{f},\hat{\mathbf{f}})$ is unknown, $\mathcal{D}$ can be approximated from a test set as $\hat{\mathcal{D}} = \frac{1}{N} \sum_{i = 1}^N d(\mathbf{f}(i),\hat{\mathbf{f}}(i))$.}, where we define the squared-error between two n-dimensional vectors $\mathbf{a}$ and $\mathbf{b}$ as $d(\mathbf{a},\mathbf{b}) \triangleq \frac{1}{n} ||\mathbf{a} - \mathbf{b}||_2^2$, and $\mathbb{E}[\cdot]$ denotes the expectation operator.
\begin{subequations}
\begin{align}
\mathcal{R} &= \frac{1}{n} \mathbb{E}[\text{\# bits used}] \label{subeq:R} \\
\mathcal{D} &= \mathbb{E}[d(\mathbf{F},\hat{\mathbf{F}})] \label{subeq:D}
\end{align}
\end{subequations}

Within the similarity search domain, for many applications, a noisy query $\mathbf{q}$ is introduced and it is desired to find $\mathcal{L}(\mathbf{q}) = \{ 1 \leqslant i \leqslant N| d(\mathbf{f}(i) , \mathbf{q} ) \leqslant \epsilon n \}$, a list of most similar items to $\mathbf{q}$ among $\mathrm{F}$. Since for large-scale problems, $\mathrm{F}$ is not available in memory and also direct matching with it is computationally expensive, an approximative $\hat{\mathcal{L}}(\mathbf{q})$ is preferred using low-complexity matching of $\mathbf{x}(i) = \mathbb{Q}[\mathbf{f}(i)]$ vs. $\mathbf{y} = \mathbb{Q}[\mathbf{q}]$, rather than $\mathbf{f}(i)$ vs. $\mathbf{q}$. So we require $\mathbb{Q}[\cdot]$ to be additionally compatible with this framework. 

The efficiency of the decoder $\mathbb{Q}^{-1}[\cdot]$ can be doubly important for search applications since we can re-order and prune the inaccurate $\hat{\mathcal{L}}(\mathbf{q})$ by reconstructing $\mathbf{f}(i)$'s with $i \in \hat{\mathcal{L}}(\mathbf{q})$ and directly match them with $\mathbf{q}$ based on $d(\hat{\mathbf{f}}(i),\mathbf{q})$.  

While this idea of ``list-refinement'' has not been particularly emphasized in the literature, lots of algorithms for fast search directly target distortion minimization as their main objective. Among the very broad literature,
not to mention the family of VQ-based methods like PQ \cite{jegou2011product} and OPQ \cite{6678503} which are codebook-based, many examples from the family of binary hashing methods also aim at distortion minimization. We can mention, e.g., the successful ITQ \cite{6296665}, which iteratively learns a projector matrix to minimize the distortion of the projected data and the corresponding binary codes. Another attempt is the Sparse Projections \cite{7298954}, an extension of ITQ for higher rates using similar objective. 

\section{Proposed: Reconstruction from STC} \label{STC_Rec}
We first review the basic STC framework for fast search in section \ref{sub:STC_ANN}. Based on its specifications, we design the single-layer reconstruction from the STC in section \ref{sub:STC_shallow} to its best. We then extend it to ML-STC, the multi-layer version to achieve near optimal distortion for all rate-regimes.
\subsection{STC framework for fast search}  \label{sub:STC_ANN}
The STC consists of a projection step\footnote{Throughout this paper, we assume that the projection matrix $\mathrm{A}$ is square and hence the dimensions of $\mathbf{f}$ and $\mathbf{x}$ are the same.}, followed by a ternary quantization. More formally, the corresponding STC for $\mathbf{f}$, i.e.,  $\mathbf{x} = \mathbb{Q}_{\text{STC}}[\mathbf{f}]$ with threshold $\lambda_X$ is: 
\begin{equation}
\mathbf{x} = \phi_{\lambda_X}(\mathrm{A}\mathbf{f}) \odot \boldsymbol{\beta}, \label{eq:STC_formulation}
\end{equation}
where $\phi_{\lambda}(x) = \text{sign}(x) \cdot \mathbbm{1}_{\lbrace |x| > \lambda \rbrace}$ is the element-wise ternary thresholding operator, `$\odot$' is the Hadamard product and $\boldsymbol{\beta} = [\beta_1, \cdots, \beta_n]^T$ is a re-weighting vector which is independent of a particular $\mathbf{f}$ and is fixed for all database vectors\footnote{In \cite{Sohrab_ISIT2017} and \cite{Sohrab_WIFS2016}, reconstruction was not considered and we had $\boldsymbol{\beta} = \mathbf{1}$.}. Therefore, each element of $\mathbf{x}$, i.e., $x_i$ has a ternary alphabet $\mathcal{X}_i = \{\pm \beta_i, 0 \}$. In practice, $\lambda_X$ is chosen such that $\mathbf{x}$ is sparse. The motivation behind such an encoding comes from similarity search where the memory and computational complexity requirements encourage sparsity and a fixed-point alphabet. 

After the enrollment of all $\mathbf{x}(i)$'s (usually stored in look-up-tables), a query vector $\mathbf{q}$ undergoes similar encoding, i.e., $\mathbf{y} = \phi_{\lambda_Y}(\mathrm{A}\mathbf{q})$ and is matched with $\mathbf{x}(i)$'s to produce $\hat{\mathcal{L}}(\mathbf{q})$. This matching can be performed very fast, e.g., using fixed-point sparse matrix multiplications. 


\subsection{Single-layer architecture} \label{sub:STC_shallow}
We formulate reconstruction of STC, i.e., $\hat{\mathbf{f}} = \mathbb{Q}^{-1}_{\text{STC}}[\mathbf{x}]$ as:
\begin{equation} \label{eq:single_layer_rec}
\hat{\mathbf{f}} = \mathrm{B} \mathbf{x} = \mathrm{B} \phi_{\lambda}(\mathrm{A}\mathbf{f}) \odot \boldsymbol{\beta}.
\end{equation}

Where $\mathrm{B}$ is the reconstruction matrix that can be learned from the training data. However, in order to avoid over-fitting, the forward projection step using $\mathrm{A}$ should be imposed as a structure to help training. So we decompose as $\mathrm{B} = (\mathrm{A}^T \mathrm{A})^{-1}\mathrm{A}^T \mathrm{B'}$ and instead optimize $\mathrm{B}'$:

\begin{equation} \label{eq:learning_pseudoInv}
\mathrm{B'} = \argmin_{\mathrm{B'}} ||\mathrm{F} - (\mathrm{A}^T \mathrm{A})^{-1}\mathrm{A}^T \mathrm{B'} \mathrm{X}||_{\mathcal{F}}^2, 
\end{equation}
where $||\cdot||_{\mathcal{F}}$ is the Frobenius norm for a matrix. This can easily be re-expressed as:
\begin{align*}
\mathrm{B'} &= \argmin_{\mathrm{B'}}  || (\mathrm{A}^T \mathrm{A}) \mathrm{F} - \mathrm{A}^T \mathrm{B'} \mathrm{X}  ||_{\mathcal{F}}^2\\
 &= \argmin_{\mathrm{B'}}  \text{Tr}\Big[ (\mathrm{A} \mathrm{F} -  \mathrm{B'}\mathrm{X})^T \mathrm{A}\mathrm{A}^T (\mathrm{A} \mathrm{F} -  \mathrm{B'}\mathrm{X}) \Big]\\
 &= \argmin_{\mathrm{B'}}  \text{Tr}\Big[ -2\mathrm{A}\mathrm{A}^T\mathrm{A}\mathrm{F}\mathrm{X}^T\mathrm{B'}^T + \mathrm{B'}\mathrm{X}\mathrm{X}^T\mathrm{B'}^T\mathrm{A}\mathrm{A}^T \Big] \label{subeq:obj}.
\end{align*}

Derivating w.r.t. $\mathrm{B'}$ and equating to zero gives:
\begin{equation*} 
\mathrm{B'} = \mathrm{A} \mathrm{F} \mathrm{X}^T (\mathrm{X}\mathrm{X}^T)^{-1}.
\end{equation*}

Assuming $\mathbf{F}$ to have a covariance matrix $\mathrm{C}_F$, i.e.,  $\mathrm{C}_F = \frac{1}{n}\mathbb{E}[\mathbf{F} \mathbf{F}^T]$, we chose the projection as $\mathrm{A} = \mathrm{U}_F^T$, where $\mathrm{C}_F = \mathrm{U}_F \mathrm{\Sigma}_F \mathrm{U}_F^T$ is the eigenvalue decomposition of $\mathrm{C}_F$.

Therefore, the projected data $\tilde{\mathbf{x}} \triangleq \mathrm{A} \mathbf{f}$  is de-correlated as $\tilde{\mathbf{X}} \sim \mathcal{N}(\mathbf{0}, \mathrm{\Sigma}_F)$\footnote{Gaussianity assumption in the projected domain is justified from CLT, even if the original data is highly non-Gaussian.}, where $\mathrm{\Sigma}_F = \text{diag} \big([\sigma_1^2, \cdots, \sigma_n^2]^T\big)$ with $\sigma_i^2$'s being the eigenvalues of $\mathrm{C}_F$ which are decaying in value for the correlated $\mathbf{F}$.
 
In our experiments, it turns out that with this choice of $\mathrm{A}$, and the optimal re-weighting vector $\boldsymbol{\beta}$ for $\mathbf{x}$ which will be described shortly,  $\mathrm{B'}$ indeed converges to the identity matrix as $N$, the number of training samples grows larger. This means that it suffices to choose $\mathrm{B'} = \mathbb{I}_n$. Equivalently stated, $\mathrm{B} = \mathrm{A}^T = \mathrm{U}_F$ would be the optimal back-projector of $\mathbf{x}$ to $\mathbf{f}$ under this setup. 

We can characterize the expected distortion for reconstruction of a random vector $\mathbf{F}$ from $\mathbf{X}$. Emphasizing the orthonormality of $\mathrm{A}$, we can then write:
\begin{align*}
\mathcal{D} &= \mathbb{E} \big[ d(\mathbf{F}, \hat{\mathbf{F}}) \big] = \frac{1}{n}\mathbb{E} \big[ ||\mathbf{F} - \mathrm{A}^T \mathbf{X}||_2^2  \big] \\
 &=  \frac{1}{n} \mathbb{E} \big[ ||\mathrm{A}\mathbf{F} - \mathbf{X})||_2^2  \big] = \frac{1}{n}\mathbb{E} \big[ ||\tilde{\mathbf{X}} - \phi_{\lambda}(\tilde{\mathbf{X}}) \odot \boldsymbol{\beta}||_2^2 \big].
 \label{subeq:Distortion}
\end{align*}
This links the distortion in the original domain with that of the projection domain. Now we should find the optimal re-weighting vector $\boldsymbol{\beta}$. 

We had that $\tilde{X}_i$'s, the elements of $\tilde{\mathbf{X}} = [\tilde{X}_1, \cdots, \tilde{X}_n]^T$ are distributed as $\tilde{X}_i \sim p(\tilde{x}_i) =\mathcal{N}(0,\sigma_i^2)$. The total distortion $\mathcal{D}$ is the sum of the distortions at each dimension as $\mathcal{D} = \sum_{i = 1}^n D_i$, which can then be written as: 
\begin{align*}
D_i &= \mathbb{E} \big[ (\tilde{X}_i - \beta_i \phi_{\lambda}(\tilde{X}_i))^2 \big]  \\
    &=  \int_{-\infty}^{-\lambda} (\tilde{x}_i + \beta_i)^2 p(\tilde{x}_i) d\tilde{x}_i + \int_{-\lambda}^{+\lambda} \tilde{x}_i^2 p(\tilde{x}_i) d\tilde{x}_i + \\
    &\int_{+\lambda}^{+\infty} (\tilde{x}_i - \beta_i)^2 p(\tilde{x}_i) d\tilde{x}_i,
\end{align*}
This integration leads to the expression of distortion as:
\begin{equation}
D_i = \sigma_i^2 + 2\beta_i^2 \mathcal{Q}\Big(\frac{\lambda}{\sigma_i}\Big) - \frac{4\beta_i \sigma_i}{\sqrt{2\pi}} \exp\Big(\frac{-\lambda^2}{2\sigma_i^2}\Big), \label{eq:Dist_per_dim}
\end{equation}
where $\mathcal{Q}(\cdot)$ is the q-function, because of which $\beta_i$ cannot be derived in closed-form. However, for a given $\sigma_i^2$ and $\lambda$, the optimal $\beta_i^*$, i.e., 
\begin{equation}
\beta_i^* = \argmin_{\beta_i} D_i, \label{eq:beta}
\end{equation} 
can easily be obtained using a simple grid-search on (\ref{eq:Dist_per_dim}).

As a summary of the single-layer reconstruction from STC, first for the encoding, $\mathrm{A}$ is chosen as the eigenvectors of $\mathrm{C}_F$ which de-correlates the projected data $\tilde{\mathbf{x}}$. The ternarization is then  performed according to (\ref{eq:STC_formulation}), for which the elements of $\boldsymbol{\beta}$ are derived according to Eqs. \ref{eq:Dist_per_dim} \& \ref{eq:beta}. For decoding , the reconstruction is done by (\ref{eq:single_layer_rec}), where we showed that $\mathrm{B} = \mathrm{A}^T$ is the optimal choice.

Having calculated $\mathcal{D}$ as a function of $\mathrm{C}_F$ and $\lambda$, we now derive $\mathcal{R}$ using the ternary entropy $H_t(\cdot)$ as:
\begin{equation}
\begin{aligned}
&\mathcal{R} =  \frac{1}{n} H_t(\mathbf{X}) = \frac{1}{n} \sum_{i = 1}^n H_t(X_i) = \\ 
 &-\frac{1}{n} \sum_{i = 1}^n \Big(2\alpha_i \log_2(\alpha_i) + (1 - 2\alpha_i) \log_2 (1 - 2\alpha_i) \Big),   
\end{aligned} \label{eq:Rate_STC}  
\end{equation}
which follows from the fact that $\tilde{\mathbf{X}}$ and hence $\mathbf{X}$ are de-correlated and hence we can assume their approximative independence. For a ternary random variable $X_i$, $\alpha_i$ is defined as $\alpha_i = \mathbb{P}[X_i = +\beta_i] = \mathbb{P}[X_i = -\beta_i]$ and completely characterizes the ternary entropy $H_t(\cdot)$. For the above setup, this can be calculated for every $X_i$, simply as $\alpha_i = \mathcal{Q}\big(\frac{\lambda}{\sigma_i}\big)$.

\begin{figure}  [!h]
   \begin{center} 
\subcaptionbox{\textit{i.i.d.}\label{subfig:SingleLayer_D_R_low}} {\includegraphics[width=0.15\textwidth]{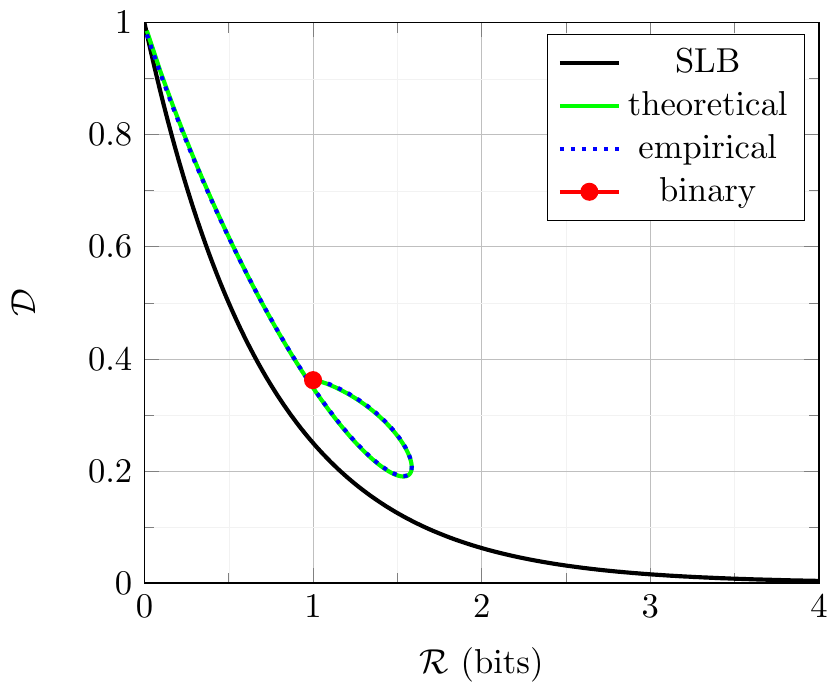}}
\subcaptionbox{mid-correlation)\label{subfig:SingleLayer_D_R_mid}} {\includegraphics[width=0.15\textwidth]{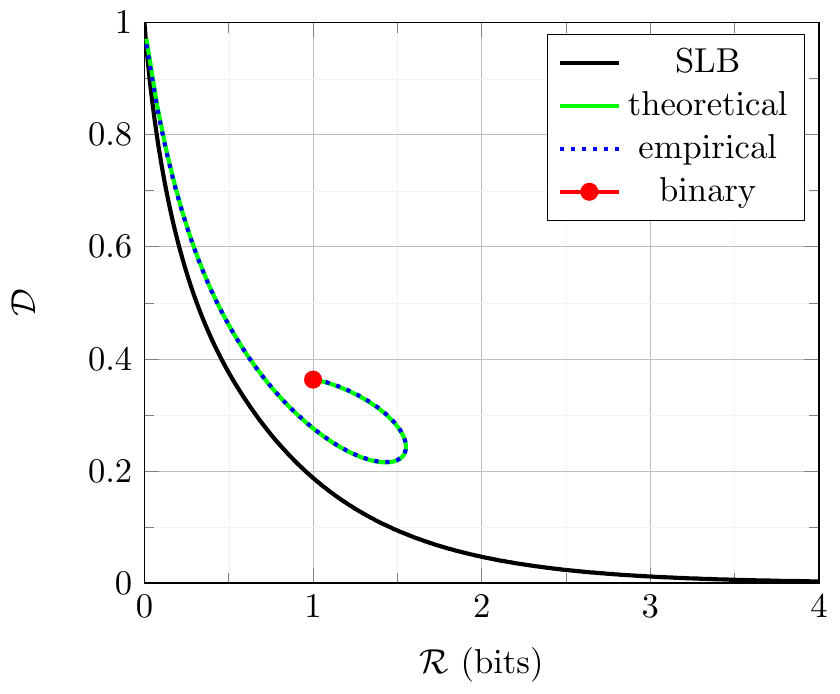}}
\subcaptionbox{high-correlation\label{subfig:SingleLayer_D_R_high}} {\includegraphics[width=0.15\textwidth]{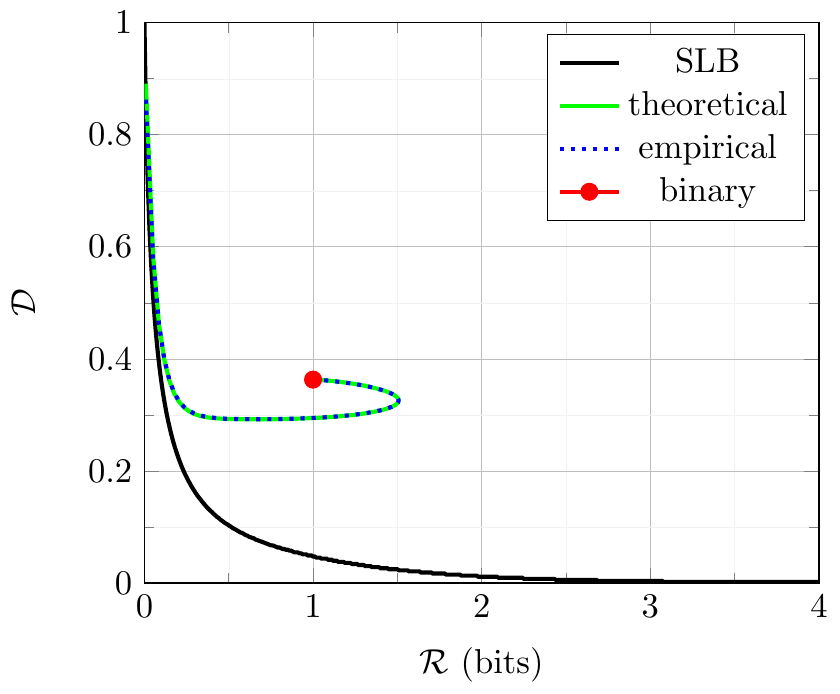}}

   \end{center}
\vspace{-0.5cm}    
   \caption{Distortion-rate curves for single-layer STC (The `theoretical' and `empirical' curves coincide very closely.)}
   \label{fig:SingleLayer_D_R}
   \end{figure}

Fig. \ref{fig:SingleLayer_D_R} shows the rate-distortion behavior of the single-layer STC for 3 different sources: (a) \textit{i.i.d.}, (b) AR(1) with $\rho = 0.5$ corresponding to mildly-correlated signals and (c) AR(1) with $\rho = 0.9$ corresponding to highly-correlated signals. For every figure, three curves are shown: the Shannon Lower Bound (SLB) derived from (\ref{eq:water-fill_distortion}) which is the theoretical lower bound achieved in the asymptotic case of $n \rightarrow \infty$ for any lossy source-coding scheme, the theoretical characterization of the STC distortion derived from (\ref{eq:Dist_per_dim}) and the empirical distortion calculated from simulations performed on $N = 10,000$ vectors of dimension $n = 500$ generated randomly. Also the case of binary encoding, i.e., zero sparsity, corresponding to $\mathcal{R} = 1$ is marked.

We clearly see that at lower sparsity levels (including the binary case) corresponding to higher rate-regimes, the single-layer structure has very poor performance. This phenomenon is due to sub-optimal rate allocation as we will describe next.
\subsubsection{Optimality of rate-allocation}
\begin{figure}  [!h]
   \begin{center}
    
\subcaptionbox{low-rate\label{subfig:RateAllocMismatch_D_R_lowRate}} {\includegraphics[width=0.15\textwidth]{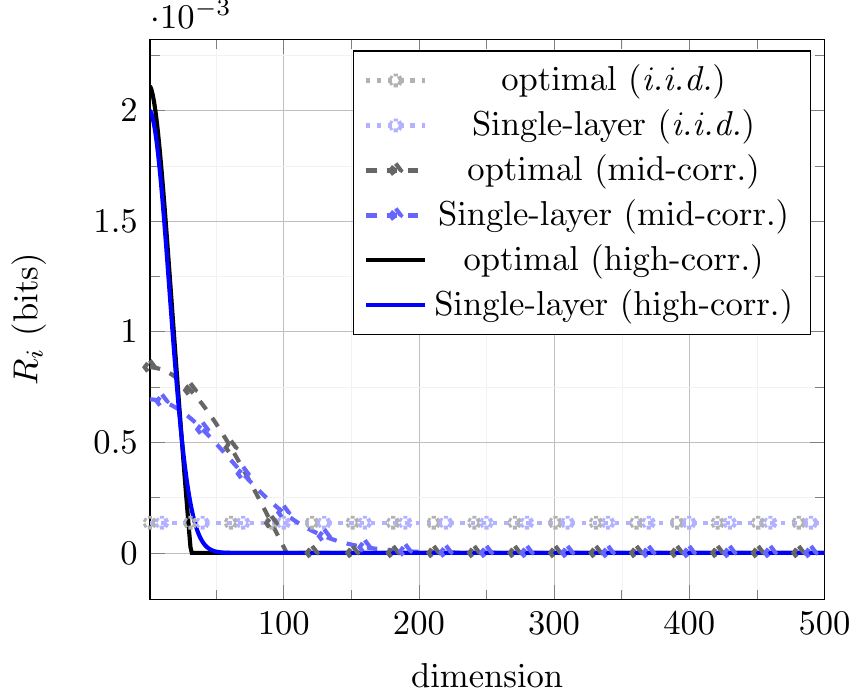}}
\subcaptionbox{mid-rate\label{subfig:RateAllocMismatch_D_R_midRate}} {\includegraphics[width=0.15\textwidth]{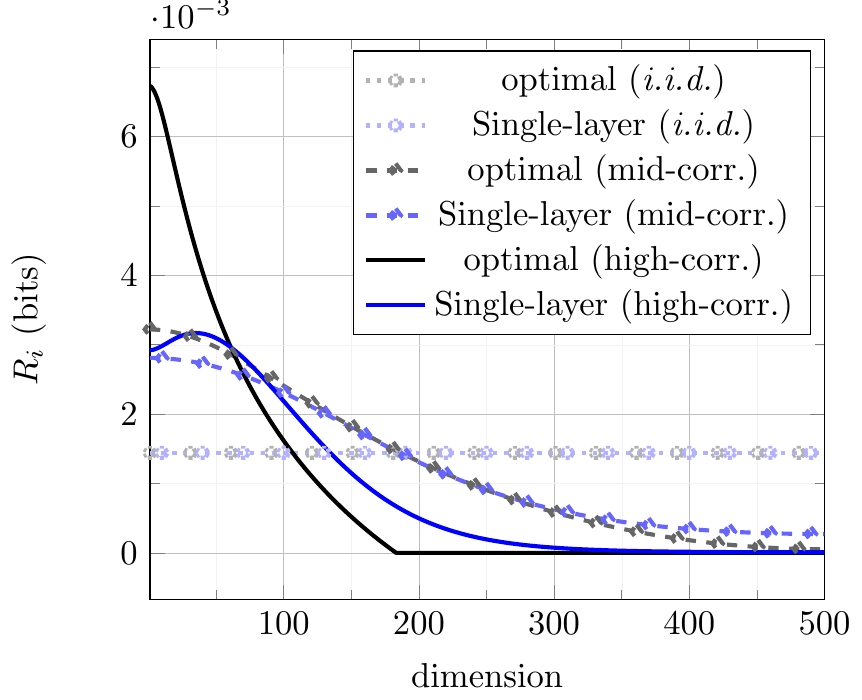}}
\subcaptionbox{high-rate\label{subfig:RateAllocMismatch_D_R_highRate}} {\includegraphics[width=0.15\textwidth]{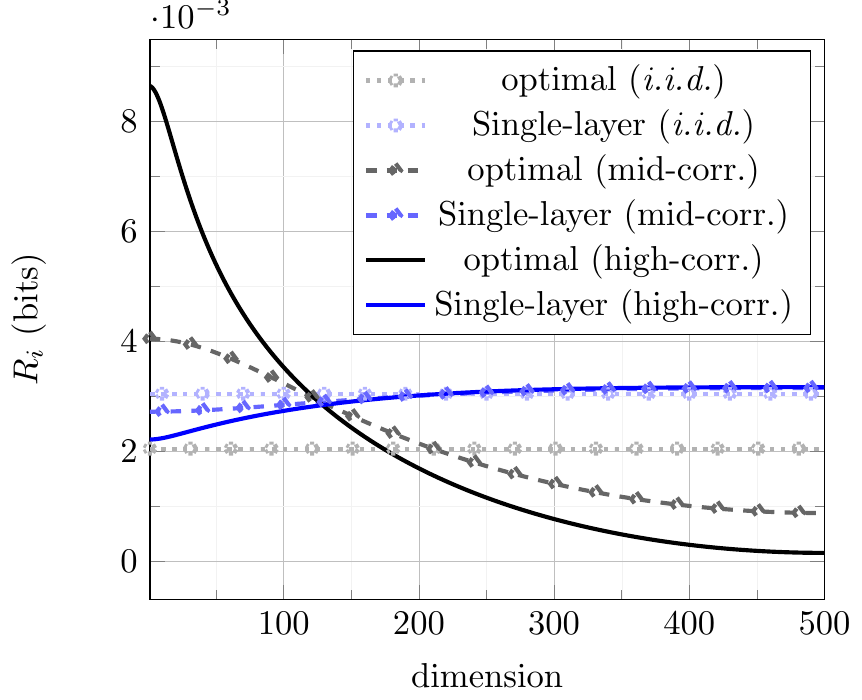}}

   \end{center}
\vspace{-0.5cm}    
   \caption{Rate allocation of single-layer STC compared to the optimal rule, under three different rate regimes and for three different sources (same as in Fig. \ref{fig:SingleLayer_D_R}).}
   \label{fig:SingleLayer_Rate_Mismatch}
   \end{figure}

The Shannon theory characterized the optimal rate allocation for $\tilde{\mathbf{X}} \sim \mathcal{N}\big(\mathbf{0},\mathrm{\Sigma}\big)$ (see Ch.10 of \cite{CoverThomas200607}). For a given $\mathcal{D}$, this is achieved when $R_i = \frac{1}{2} \log_2 \big(\frac{\sigma_i^2}{D_i}\big)$ bits are allocated for each $\tilde{X}_i$, where $D_i$, corresponding distortion of each $\tilde{X}_i$ is:
\begin{equation} \label{eq:water-fill_distortion}
D_i =
\begin{cases}
   \lambda ,& \text{if   } \sigma_i^2 \geqslant \lambda \\
    \sigma_i^2, & \text{if   } \sigma_i^2 < \lambda,
\end{cases}
\end{equation} 
and $\lambda$ is chosen such that $\sum_{i=1}^n D_i = \mathcal{D}$. The total rate is then calculated as: 
\begin{equation}
\mathcal{R}(\mathcal{D}) = \sum_{i = 1}^n R_i = \sum_{i = 1}^n \frac{1}{2} \log_2 \Big(\frac{\sigma_i^2}{D_i}\Big). \label{eq:optimal_total_rate}
\end{equation}
Comparing this optimal rate allocation of (\ref{eq:optimal_total_rate}) with the single-layer STC of (\ref{eq:Rate_STC}) reveals the fact that while they closely approximate the optimal rule at low rates, single-layer structure largely deviates from the optimal allocation at higher rates. This explains the saturating behavior of the rate-distortion curve at Fig. \ref{fig:SingleLayer_D_R}. This phenomenon is illustrated in Fig. \ref{fig:SingleLayer_Rate_Mismatch}.


\subsection{Multi-layer architecture} \label{sub:STC_deep}
\begin{figure}  [!h]
   \begin{center} 
\subcaptionbox{low-correlation\label{subfig:MultiLayer_D_R_low}} {\includegraphics[width=0.15\textwidth]{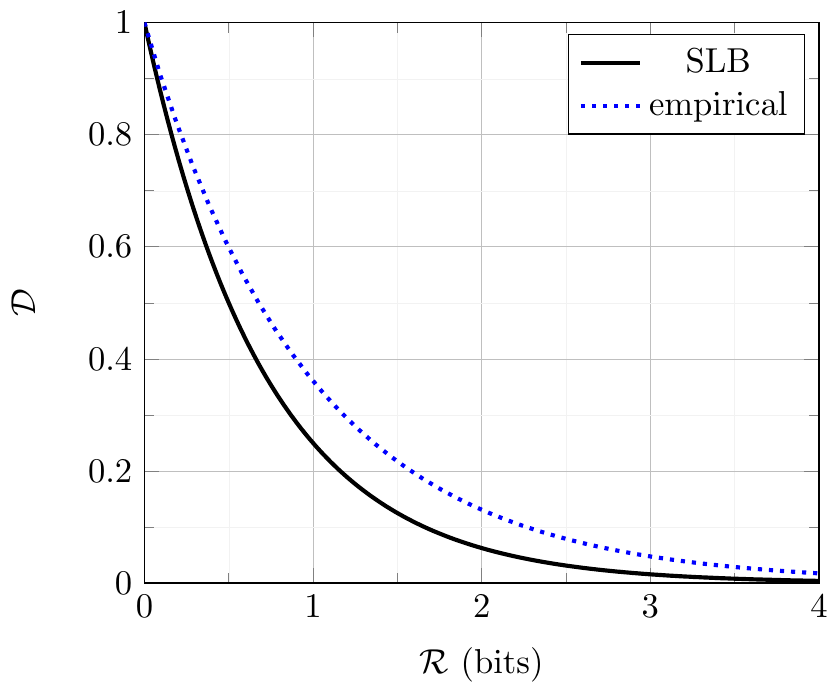}}
\subcaptionbox{mid-correlation)\label{subfig:MultiLayer_D_R_mid}} {\includegraphics[width=0.15\textwidth]{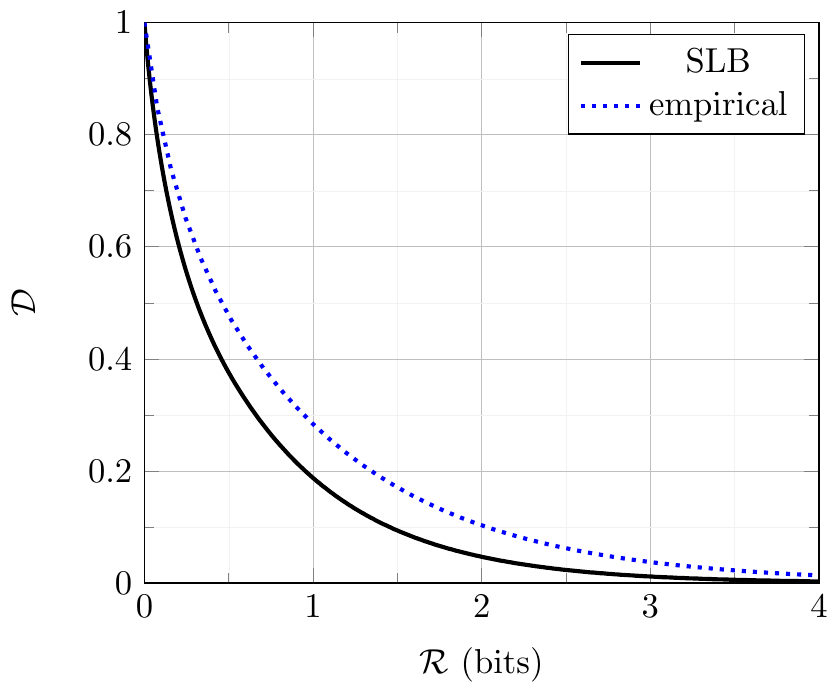}}
\subcaptionbox{high-correlation\label{subfig:MultiLayer_D_R_high}} {\includegraphics[width=0.15\textwidth]{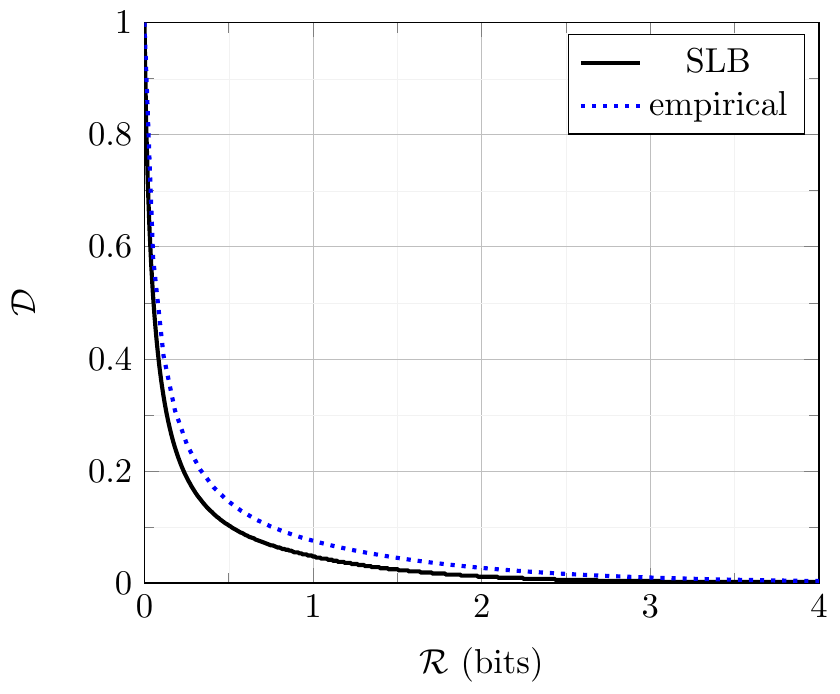}}

   \end{center}
\vspace{-0.5cm}    
   \caption{Rate distortion performance of multi-layer STC}
   \label{fig:Multilayer_D_R}
   \end{figure}

The mismatch between rate-allocation of the single-layer STC and the optimal rule at high rates limits their efficient use only for very low rate regimes.  One could think of more intricate quantization schemes than ternary to overcome this issue. However, this would ruin their simplicity and usefulness for many applications like fast similarity search. Instead, we keep operating STC at lower rates but use them in succession. This idea is demonstrated in (\ref{eq:ML-STC}):
\begin{equation}
\begin{aligned}
\mathbf{x}^{[l]} &= \phi_{\lambda_X}^{[l]}(\mathrm{A}^{[l]} \mathbf{f}^{[l-1]}) \odot \boldsymbol{\beta}^{[l]}, \\
\mathbf{f}^{[l]} &= \mathbf{f}^{[l-1]} - \mathrm{B}^{[l]}  \mathbf{x}^{[l]}.  
\end{aligned} \label{eq:ML-STC}
\end{equation}
The superscripts depict the index of the layer $l = 1, \cdots, L$. $\mathbf{f}^{[l]}$is the input to the algorithm at layer $l$ which is the residual of the approximation from layer $l-1$ and is initialized as $\mathbf{f}^{[0]} = \mathbf{f}$. The rest of the procedure is the same as the single-layer case. Fig. \ref{fig:Multilayer_D_R} shows the success of this idea in rate-distortion within the same setup as Fig. \ref{fig:SingleLayer_D_R}. 

\section{Experiments}
\label{exp}
\begin{figure}  [!h]
 \vspace{-0.4cm}
   \begin{center} 
\subcaptionbox{MNIST} {\includegraphics[width=0.22\textwidth]{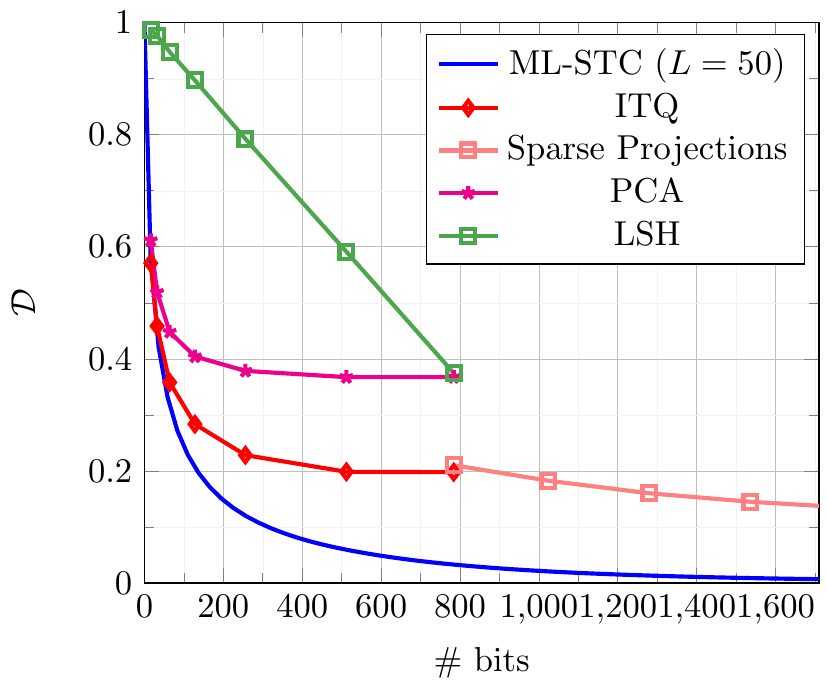}}
 \subcaptionbox{GIST-1M} {\includegraphics[width=0.22\textwidth]{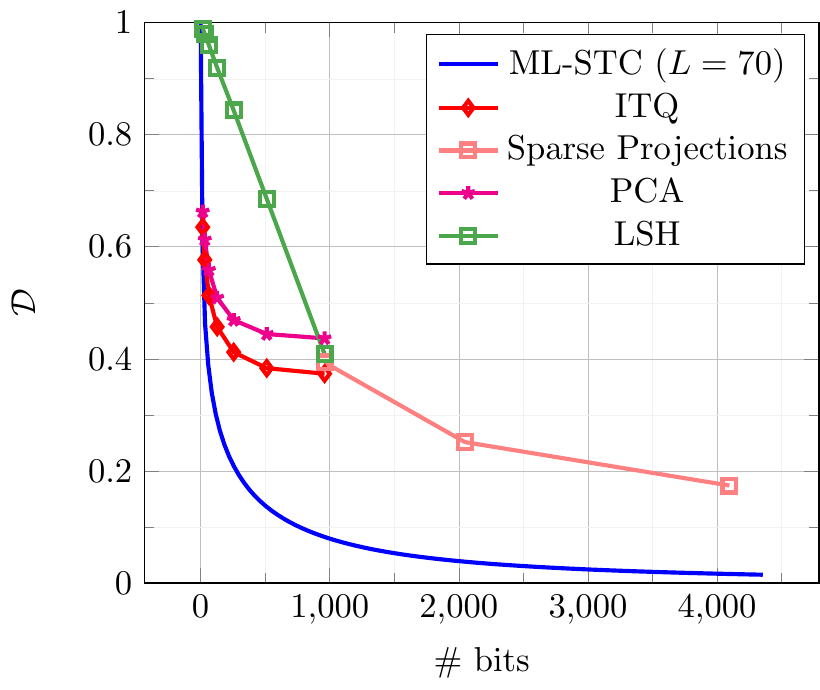}}
   \end{center}
\vspace{-0.75cm}
\caption{Reconstruction distortion from compressed ANN sets.}    
   \label{fig:DatabaseCompression}
   \end{figure}

Here we demonstrate the performance of the proposed algorithm on the task of database compression. We use two public databases, MNIST of mid-scale and the large-scale GIST-1M set \cite{jegou2011product}. The MNIST contains $60,000$ train and $10,000$ test images with $784$ pixels which we consider as feature vectors. The GIST-1M comprises of $960$-dimensional GIST descriptors with $500,000$ train and $1$ million test vectors. Along with our ML-STC, we also experiment with the ITQ \cite{6296665}, the Sparse Projections \cite{7298954} (using sparsity $=50\%$), PCA hashing and the LSH (Sim-Hash) \cite{charikar2002similarity}. We train all algorithms on the train and calculate the distortion on the test set. The reconstruction from the binary codes consists of pseudo-inversion and, for PCA hashing, the above-mentioned re-weighting stage followed by the inversion. For the ITQ and Sparse Projections, the vector $\boldsymbol{\beta}$ is irrelevant and also detrimental according to their objective functions. Instead, a scaler-valued optimal $\beta$ is learned from the training set as $\beta \!=\! \frac{\text{Tr}[\mathrm{F}\hat{\mathrm{F}}^T]}{\text{Tr}[\hat{\mathrm{F}} \hat{\mathrm{F}}^T]}$ and multiplied globally as $\hat{\mathrm{F}} \!\leftarrow\! \beta \hat{\mathrm{F}}$. Fig. \ref{fig:DatabaseCompression} sketches the results of these experiments. The ML-STC outperforms others with a large margin.


\section{Conclusions}
\label{summary}
A universal compressor network is designed based on the Sparse Ternary Codes framework for similarity search where we demonstrate rate-distortion performance on synthetic as well as real data, superior to several exemplar methods from the binary encoding family. The intrinsic limitations of rate allocation w.r.t. optimality suggests a multi-layer design which is hopeless for binary encoding but applies very nicely for STC when they are set to be highly sparse. Thanks to the simplicity of encoding and the universality of the signals considered, these results can be useful for many applications. An immediate benefit would be for the idea of list-refinement in similarity search which we will address in a future work.

\clearpage
\bibliographystyle{IEEEbib}

\bibliography{Bibliography}

\end{document}